\documentclass[epsf,onecolumn,11pt]{IEEEtran}
\usepackage{graphicx}

\def\be{\begin{eqnarray}}
\def\en{\end{eqnarray}}
\def\bee{\begin{eqnarray*}}
\def\enn{\end{eqnarray*}}

\newtheorem{remark}{Remark}[section]

\begin{document}

\title{A Family of Simplified Geometric Distortion Models for Camera Calibration
\thanks{All correspondence should be addressed to Dr. YangQuan Chen. Tel.: 1(435)797-0148, Fax: 1(435)797-3054,  Email: \texttt{yqchen@.ece.usu} \texttt{.edu}. CSOIS URL: \texttt{http://www.csois.usu.edu/}}
}

\author{Lili Ma, {\it Student Member, IEEE}, YangQuan Chen and Kevin L. Moore, {\it Senior Members, IEEE}
\\ Center for Self-Organizing and Intelligent Systems (CSOIS),\\Dept. of Electrical and Computer Engineering, 4160 Old Main Hill,\\ Utah State University (USU), Logan, UT 84322-4160, USA.\\}
\maketitle{}

\begin{abstract}
The commonly used radial distortion model for camera calibration is in fact an assumption or a restriction. 
In practice, camera distortion could happen in a general geometrical manner that is not limited to the radial sense. 
This paper proposes a simplified geometrical distortion modeling method by using two different radial distortion functions in the two image axes.
A family of simplified geometric distortion models is proposed, which are either simple polynomials or the rational functions of polynomials. 
Analytical geometric undistortion is possible using two of the distortion functions discussed in this paper and their performance can be improved by applying a piecewise fitting idea. 
Our experimental results show that the geometrical distortion models always perform better than their radial distortion counterparts. 
Furthermore, the proposed geometric modeling method is more appropriate for cameras whose distortion is not perfectly radially symmetric around the center of distortion. 
\end{abstract}

\section{Introduction}
For many computer vision applications, such as robot visual inspection and industrial metrology, where a camera is used as a sensor in the system, the camera is usually assumed to be fully calibrated beforehand. 
Camera calibration is the estimation of a set of parameters that describes the camera's imaging process. With this set of parameters, a perspective projection matrix can directly link a point in the 3-D world reference frame to its projection (undistorted) on the image plane. This is given by:
\begin{equation}
\label{eqn: projection matrix}
\lambda \left [\matrix{u \cr v \cr 1} \right ] 
= {\bf A} \, \left[{\bf R} \mid {\bf t}\right] \left [\matrix{X^w \cr Y^w \cr Z^w \cr 1} \right ] 
= \left[\matrix{\alpha & \gamma &u_0\cr 0 & \beta & v_0\cr 0 & 0 &1}\right] \left [\matrix{X^c \cr Y^c \cr Z^c} \right ],
\end{equation}
where $(u,v)$ is the distortion-free image point on the image plane. The matrix $\bf A$ fully depends on the camera's 5 intrinsic parameters $(\alpha, \gamma, \beta, u_0, v_0)$, with $(\alpha, \beta)$ being two scalars in the two image axes, $(u_0, v_0)$ the coordinates of the principal point, and $\gamma$ describing the skewness of the two image axes. $[X^c, Y^c, Z^c]^T$ denotes a point in the camera frame that is related to the corresponding point $[X^w, Y^w, Z^w]^T$ in the world reference frame by $P^c = {\bf R} P^w + \bf t$, with $({\bf R}, {\bf t})$ being the rotation matrix and the translation vector. 

\begin{figure}[htb]
\centering
\includegraphics[width=0.8\textwidth]{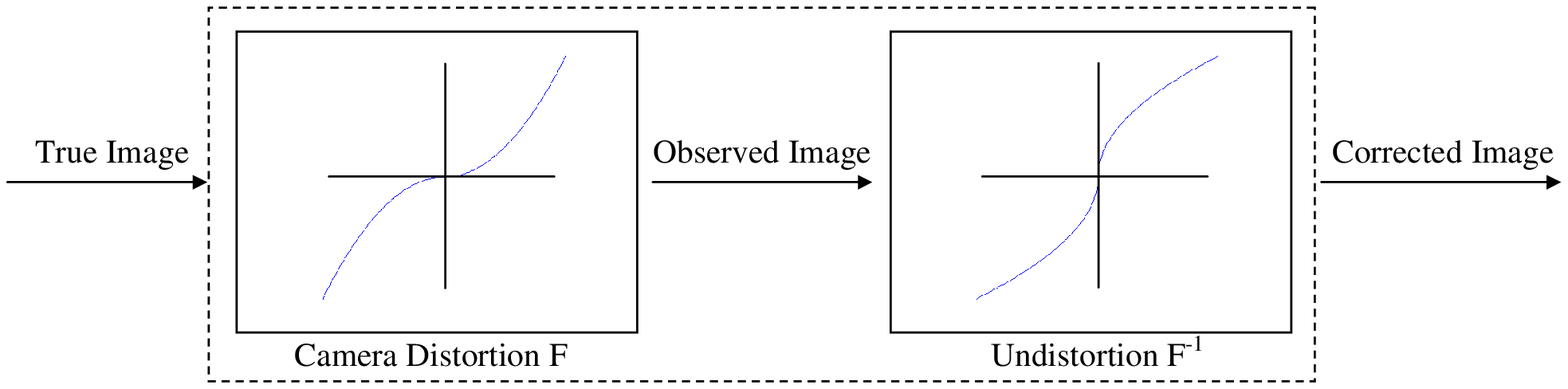}
\caption{Lens distortion and undistortion.}
\label{fig: one-to-one}
\end{figure}

In camera calibration, lens distortion is very important for accurate 3-D measurement \cite{Tsai88Techniques}. The lens distortion introduces certain amount of nonlinear distortions, denoted by a function $F$ in Fig.~\ref{fig: one-to-one}, to the true image. The observed distorted image thus needs to go through the inverse function $F^{-1}$ to output the corrected image. That is, the goal of lens undistortion, or image correction, is to achieve an overall one-to-one mapping.

Among various nonlinear distortions, the radial distortion, which is along the radial direction from the center of distortion, is the most severe part \cite {OlivierF01Straight,tsai87AVersatile}. 
The removal or alleviation of radial distortion is commonly performed by first applying a parametric radial distortion model, estimating the distortion coefficients, and then correcting the distortion. 
Most of the existing works on the radial distortion models can be traced back to an early study in photogrammetry \cite{Photogrammetry80}, where the radial distortion is governed by the following polynomial equation \cite{Photogrammetry80,zhang99calibrationinpaper,Heikkil97fourstepcameracalibration,Janne96Calibration}:
\begin{equation}
\label{eqn: general polynomial}
r_d = r + \delta_r = r \, f(r, {\bf k}) = r \, (1 + k_1 r^2 + k_2 r^4 + k_3 r^6+ \cdots),
\end{equation}
which is equivalent to
\begin{equation}
\label{eqn: radial}
x_d = x \,f(r, {\bf k}), \;\; y_d = y \,f(r, {\bf k}),
\end{equation}
where ${\bf k} = [k_1, k_2, k_3, \ldots]$ is a vector of distortion coefficients. 

For cameras whose distortion is not perfectly radially symmetric around the center of distortion (which is assumed to be at the principal point in our discussion), radial distortion modeling will not be accurate enough for applications such as precise visual metrology. In this case, a more general distortion model, i.e., geometric distortion, needs to be considered. 
In this work, a simplified geometric distortion modeling method is proposed, where two different functions in the form of a variety of polynomial and rational functions are used to model the distortions along the two image axes. 
The proposed simplified geometric distortion models are simpler in structure than that considered in \cite{Janne00Geometric} where the total geometric distortion consists of the radial distortion and the decentering distortion. The geometric distortion modeling method proposed is a lumped distortion model that includes all the nonlinear distortion effects.

For real time image processing applications, the property of having analytical undistortion formulae is a desirable feature for both the radial and the geometric distortion models. Though there are ways to approximate the undistortion without numerical iterations, having analytical inverse formulae is advantageous by giving the exact inverse without introducing extra error sources. The key contribution of this paper is the proposition of a family of simplified geometric distortion models that can achieve comparable calibration accuracy to that in \cite{Janne00Geometric} and better performance than their radial distortion modeling counterparts. For fairness, these comparisons are based on the same (or reasonable) numbers of distortion coefficients. To preserve the property of having analytical inverse formulae with satisfactory calibration accuracy, a piecewise fitting idea is applied to the simplified geometric modeling for two particular rational distortion functions presented in Sec.~\ref{sec: polynomial and rational models}. 

The rest of the paper is organized as follows. Sec.~\ref{sec: polynomial and rational models} summarizes some existing polynomial and rational radial distortion models that can also be applied to model the geometric distortion. The simplified geometric distortion modeling method is proposed in Sec.~\ref{sec: geometric}. Experimental results and comparisons between the simplified geometric and the radial distortion models are illustrated and discussed in Sec.~\ref{sec: experimental results}. Finally, some concluding remarks are given in Sec.~\ref{sec: conclusion}. 
The variables used throughout this paper are listed in Table~\ref{table: variables used}. 

\begin{table}[htb]
\centering
\caption{List of Variables}
\label{table: variables used}
\renewcommand{\arraystretch}{1.3}
\vspace{-2mm}
{\small
{\begin {tabular}{|c|l|}\hline
{\bf Variable} & {\bf Description} \\[1ex]\hline
$(u_d, \, v_d)$              & Distorted image point in pixel\\\hline
$(u, \, v)$                  & Distortion-free image point in pixel\\\hline
$(x_d, \, y_d)$              & $[x_d, \, y_d, \, 1]^T = A^{-1} [u_d, \, v_d, \, 1]^T$\\\hline
$(x, \, y)$                  & $[x,\, y,\, 1]^T = A^{-1} [u,\, v,\, 1]^T$ \\\hline
$r_d$                        & $r_d^2 = x_d^2 + y_d^2$ \\\hline
$r$                          & $r^2 = x^2 + y^2$ \\\hline
$\bf k$                      & Distortion coefficients (radial or geometric) \\\hline
\end {tabular}}}   
\end{table}

\section{Polynomial and Rational Distortion Functions}
\label{sec: polynomial and rational models}

The commonly used polynomial radial distortion model is given in the form of (\ref{eqn: general polynomial}). In this paper, we consider both the polynomial (functions $\# \, 1,2,3$ in Table~\ref{table: 10 models}) and rational radial distortion functions (functions $\# \, 5,6,7,8,9,10$ in Table~\ref{table: 10 models}) \cite{Book00MulGeo,LiliSub2ITMI03Rational}. Clearly, all these functions in Table~\ref{table: 10 models}, except the function $\#4$, are special cases of the following radial distortion function having analytical inverse formulae:
\begin{equation}
\label{eqn: general analytical}
f(r, {\bf \kappa}) = \frac{1 + \kappa_1 \,r + \kappa_2 \,r^2}{1 + \kappa_3 \,r + \kappa_4 \,r^2 + \kappa_5 \,r^3}.
\end{equation}
For example, when $\kappa_1 = 0, \kappa_5 = 0$, equation (\ref{eqn: general analytical}) becomes the function $\#10$ in Table~\ref{table: 10 models} with 
$k_1 = \kappa_2$, $k_2 = \kappa_3$, and $k_3 = \kappa_4$. 
Notionwise, $k_1$, $k_2$, and $k_3$ here correspond to their specific distortion function, i.e.,  $k_1$, $k_2$, and $k_3$ do not have a global meaning.
The function $\#4$ in Table~\ref{table: 10 models}   is in the form of (\ref{eqn: general polynomial}) with 2 distortion coefficients, which is the most commonly used conventional radial distortion function in the polynomial approximation category. The other 9 functions in Table~\ref{table: 10 models} are studied specifically with the goal to achieve comparable performance with the function $\#4$ using the least amount of model complexity and as few distortion coefficients as possible. Since the functions $\#9$ and $\#10$ in Table~\ref{table: 10 models} begin to show comparable calibration performance to the function $\#4$ (as can be seen later in Table~\ref{table: SGR comparisons}) \cite{LiliSub2ITMI03Rational}, more complex distortion functions are not studied in this work. 

\begin{table}[htb]
\centering
\caption{Polynomial and Rational Distortion Functions}
\label{table: 10 models}
\renewcommand{\arraystretch}{1.15}
\vspace{-2 mm}
\setlength{\tabcolsep}{1.6mm}
{\small
{\begin {tabular}{|c|l|c|l|}\hline
$\bf \#$ & $f(r,{\bf k})$ & $\bf \#$ & $f(r,{\bf k})$\\[0.7ex]\hline
1 & $1 + k_1 \, r$                & 6 & $1/(1 + k_1 \,r^2)$                            \\\hline
2 & $1 + k_1 \, r^2$              & 7 & $(1 + k_1 \,r)/(1 + k_2 \,r^2)$              \\\hline
3 & $1 + k_1 \, r+k_2 \, r^2$     & 8 & $1/(1 + k_1 \,r + k_2 \,r^2)$                \\\hline
4 & $1 + k_1 \, r^2 + k_2 \, r^4$ & 9 & $(1 + k_1 \, r)/(1 + k_2 \, r + k_3 \, r^2)$ \\\hline
5 & $1/(1 + k_1 \,r)$             &10 & $(1 + k_1 \,r^2)/(1 + k_2 \,r + k_3 \,r^2)$  \\\hline
\end {tabular}}}
\end{table}

Notice that all the functions in Table~\ref{table: 10 models} satisfy the following properties: 
\begin{itemize}
\item [\bf 1)] The function is radially symmetric around the center of distortion and is expressed in terms of the radius $r$ only; 
\item [\bf 2)] The function is continuous and $r_d = 0$ iff $r = 0$;
\item [\bf 3)] The approximation of $x_d$ is an odd function of $x$. 
\end{itemize}
The above three properties act as the criteria to be a candidate for the radial distortion function. However, for the general geometric distortion functions, which are not necessarily the same along the two image axes, the first property does not need to be satisfied, though the functions need to be continuous such that there will be no distortion only at the center of distortion. 

The well-known radial distortion model (\ref{eqn: general polynomial}) that describes the laws governing the radial distortion does not involve a quadratic term. 
Thus, it might be unexpected to add one. However, when interpreting from the relationship between $(x_d,y_d)$ and $(x,y)$ in the camera frame, the purpose of radial distortion function is to approximate the $x_d \leftrightarrow x$ relationship, which is intuitively an odd function. Adding a quadratic term to $\delta_r$ does not alter this fact as shown in \cite{LiliISIC03Flex}. As demonstrated in \cite{LiliISIC03Flex}, it is reasonable to introduce a quadratic term  to $\delta_r$ to broaden the choice of radial distortion functions with a better calibration fit. Therefore, as long as the above listed three properties are satisfied, there should be no restriction in the form of $\delta_r$. With this argument in mind, we also proposed the rational radial distortion models with analytical undistortion formulae as shown in Table~\ref{table: 10 models}, with details presented in  \cite{LiliSub2ITMI03Rational}.

To compare the performance of the simplified geometric distortion models with their radial distortion counterparts, the calibration procedures presented in \cite{zhang99calibrationinpaper} are applied.
In \cite{zhang99calibrationinpaper}, the estimation of radial distortion is done after having estimated the intrinsic and the extrinsic parameters, just before the nonlinear optimization step. So, for different distortion models (radial or geometric), we can reuse the estimated intrinsic and extrinsic parameters. 
To compare the performance of different distortion models, the final value of optimization function $J$, which is defined to be \cite{zhang99calibrationinpaper}:
\begin{equation}
\label{eqn: objective function}
J = \sum_{i=1}^N \sum_{j=1}^n \|m_{i j}-\hat m({\bf A}, {\bf k}, {\bf R}_i, {\bf t}_i, M_j) \|^2,
\end{equation}
is used, where $\hat m({\bf A}, {\bf k}, {\bf R}_i, {\bf t}_i, M_j)$ is the projection of point $M_j$ in the $i^{th}$ image using the estimated parameters, $\bf k$ denotes the distortion coefficients (radial or geometric), $M_j$ is the $j^{th}$  3-D point in the world frame with $Z^w = 0$, $n$ is the number of feature points in the coplanar calibration object, and $N$ is the number of images taken for calibration.

\section{Simplified Geometric Distortion Models}
\label{sec: geometric}
\subsection{Model}

A family of simplified geometric distortion models is proposed as
\begin{equation}
\label{eqn: U-D model in xdyd}
x_d = x \,f(r, {\bf k}_1), \;\; y_d = y \,f(r, {\bf k}_2),
\end{equation}
where the distortion function $f(r, {\bf k})$ in (\ref{eqn: U-D model in xdyd}) can be chosen to be, though not restricted to, any of the functions in Table~\ref{table: 10 models}. When ${\bf k}_1 = {\bf k}_2 = {\bf k}$, the geometric distortion reduces to the radial distortion in equation (\ref{eqn: radial}). From (\ref{eqn: U-D model in xdyd}), the relationship between $(u_d,v_d)$ and $(u,v)$ becomes
\begin{equation}
\label{eqn: udvd from xdyd}
\left\{\hspace{-1mm}
\begin{array}{l}
u_d - u_0 = (u - u_0) \,f(r, {\bf k}_1) \\
\hspace{15mm} + \, \gamma/\beta \,(v - v_0) \,[f(r, {\bf k}_2) - f(r,{\bf k}_1)]\\
\hspace{0.6mm} v_d - v_0 = (v - v_0) \,f(r,{\bf k}_2)
\end{array}\right.\hspace{-1mm}.
\end{equation}
If we define
\begin{equation}
\label{eqn: U-D model in udvd}
\left\{\hspace{-1mm}
\begin{array}{c}
u_d - u_0 = (u - u_0) \,f(r, {\bf k}_1)\\
v_d - v_0 = (v - v_0) \,f(r, {\bf k}_2)
\end{array}\right.\hspace{-1 mm},
\end{equation}
the relationship between $(x_d,y_d)$ and $(x,y)$ becomes
\begin{equation}
\label{eqn: xdyd from udvd}
\left\{\hspace{-1mm}
\begin{array}{l}
x_d = x \,f(r, {\bf k}_1) + \gamma/\alpha \;y \,[f(r, {\bf k}_1) - f(r, {\bf k}_2)]\\
y_d = y \,f(r, {\bf k}_2)
\end{array}\right.\hspace{-1 mm}.
\end{equation}
After nonlinear optimization, the final values of $J$, the intrinsic and extrinsic parameters, and the distortion coefficients using the pair (\ref{eqn: U-D model in xdyd}), (\ref{eqn: udvd from xdyd}) and (\ref{eqn: U-D model in udvd}), (\ref{eqn: xdyd from udvd}) are extremely close. Thus, in this paper, we only focus on the pair (\ref{eqn: U-D model in xdyd}), (\ref{eqn: udvd from xdyd}), while being aware that similar results can be achieved using (\ref{eqn: U-D model in udvd}), (\ref{eqn: xdyd from udvd}).

\begin{remark}
The distortion models discussed in this paper belong to the category of {\underline U}ndistorted-{\underline D}istorted model, while the {\underline D}istorted-{\underline U}ndistorted model also exists in the literature to correct distortion \cite{Toru02Unified}. The idea of simplified geometric distortion modeling can be applied to the D-U formulation simply by defining
\begin{equation}
\label{eqn: D-U model in xy}
x = x_d \, f(r_d, {\tilde {\bf k}}_1),\;\; y = y_d \, f(r_d, {\tilde {\bf k}}_2).
\end{equation}
Consistent improvement can be achieved in the above D-U formulation. 
\end{remark}

In \cite{Janne00Geometric}, the geometric distortion modeling when written in the U-D formulation is presented as: 
\begin{equation}
\label{eqn: geometric using circular control points}
\left\{\hspace{-1mm}
\begin{array}{l}
u_d = {\bar u}\, (1+k_1 \,r^2 + k_2 \,r^4 + k_3 \,r^6 + \cdots) + u_0 \\
\hspace{8mm} \,+ (2 p_1 \,{\bar u} {\bar v} + p_2 \,(r^2 + 2 {\bar u}^2)) (1 + p_3 \,r^2 + \cdots) \\
v_d = {\bar v}\, (1+k_1 \,r^2 + k_2 \,r^4 + k_3 \,r^6 + \cdots) + v_0\\
\hspace{8mm} \,+ (p_1 \,(r^2 + 2 {\bar v}^2) + 2 p_2 \,{\bar u} {\bar v}) (1 + p_3 \,r^2 + \cdots)
\end{array}\right. ,
\end{equation}
where ${\bar u} = u - u_0, {\bar v} = v - v_0$. The parameters $(k_1, k_2, k_3)$ are the coefficients for the radial distortion and the parameters $(p_1, p_2, p_3)$ are for the decentering distortion. Compared with (\ref{eqn: geometric using circular control points}), the proposed simplified geometric distortion modeling (\ref{eqn: U-D model in xdyd}), (\ref{eqn: udvd from xdyd}) is simpler in the structure and it is a lumped distortion model that includes all the nonlinear distortion effects. 

\begin{remark}
The two functions that model the distortion in the two image axes are not necessarily of the same form or structure. That is, equation (\ref{eqn: U-D model in xdyd}) can be extended to have the following more general form
\begin{equation}
\label{eqn: U-D model in xdyd general}
x_d = x \,f_x(r, {\bf k}_1), \;\; y_d = y \,f_y(r, {\bf k}_2).
\end{equation}
However, since we have no prior information as to how the distortions proceed along the two image axes, model (\ref{eqn: U-D model in xdyd general}) is not investigated in this work for lack of motivation. Of course, by choosing  $f_x(r, {\bf k}_1)$ and $f_y(r, {\bf k}_2)$ differently, there is a chance to get an even better result at the expense of  making more efforts in figuring out what the best combination should be.
\end{remark}

\subsection{Geometric Undistortion}

For the simplified geometric distortion model (\ref{eqn: U-D model in xdyd}), the property of having analytical undistortion formulae is not preserved for most of the functions in Table~\ref{table: 10 models} as for the radial distortion.
However, when using the function $\#5$ and $\#6$ in Table~\ref{table: 10 models}, the geometric undistortion can be performed analytically. For example, when
\begin{equation}
\label{eqn: example}
f(r, k_1) = \frac{1}{1 + k_1 r}, \;\; f(r, k_2) = \frac{1}{1 + k_2 r},
\end{equation}
from (\ref{eqn: U-D model in xdyd}), we have
\begin{equation}
\label{eqn: xdyd 5}
x_d = x \, \frac{1}{1 + k_1 r}, \;\; y_d = y \, \frac{1}{1 + k_2 r}.
\end{equation}
The geometric undistortion problem is to calculate $(x, y)$ from $(x_d, y_d)$ given the distortion coefficients $(k_1, k_2)$ that are determined through the nonlinear optimization process. From equation (\ref{eqn: xdyd 5}), we have the following quadratic function of $r$
\begin{equation}
\label{eqn: 2nd}
x_d^2 \, (1 + k_1 \, r)^2 + y_d^2 \, (1 + k_2 \, r)^2 = r^2,
\end{equation}
whose analytical solutions exist. The above quadratic function in $r$ has two analytical solutions, where one solution can be discarded because it deviates from $r_d$ dramatically. After $r$ is derived, $(x,y)$ can be calculated from $(x_d, y_d)$ uniquely. In this way, the geometric undistortion using the function $\#5$ in Table~\ref{table: 10 models} can be achieved non-iteratively. For the function $\#6$, a similar quadratic function in the form of $x_d^2 \, (1 + k_1 \, {\bar r})^2 + y_d^2 \, (1 + k_2 \, {\bar r})^2 = {\bar r}$ can be derived with ${\bar r} = r^2$.

\subsection{Piecewise Geometric Distortion Models Using Functions $\#5$ and $\#6$ in Table~\ref{table: 10 models}}

For real time image processing applications, geometric distortion models with analytical undistortion formulae are very desirable for the exact inverse. 
When there is no analytical undistortion formula and to avoid performing the undistortion via numerical iterations, there are ways to approximate the undistortion, such as the model described in \cite{Janne00Geometric} for the radial undistortion, where $r$ can be calculated from $r_d$ by
\begin{equation}
\label{eqn: undistortion approximation}
r =  r_d \, f(r_d, -{\bf k}).
\end{equation}
The fitting results given by the above model can be satisfactory when the distortion coefficients are small values. However,   equation (\ref{eqn: undistortion approximation}) itself introduces additional error that will inevitably degrade the  overall calibration accuracy.

\begin{figure}[htb]
\centering
\includegraphics[width=0.5\textwidth]{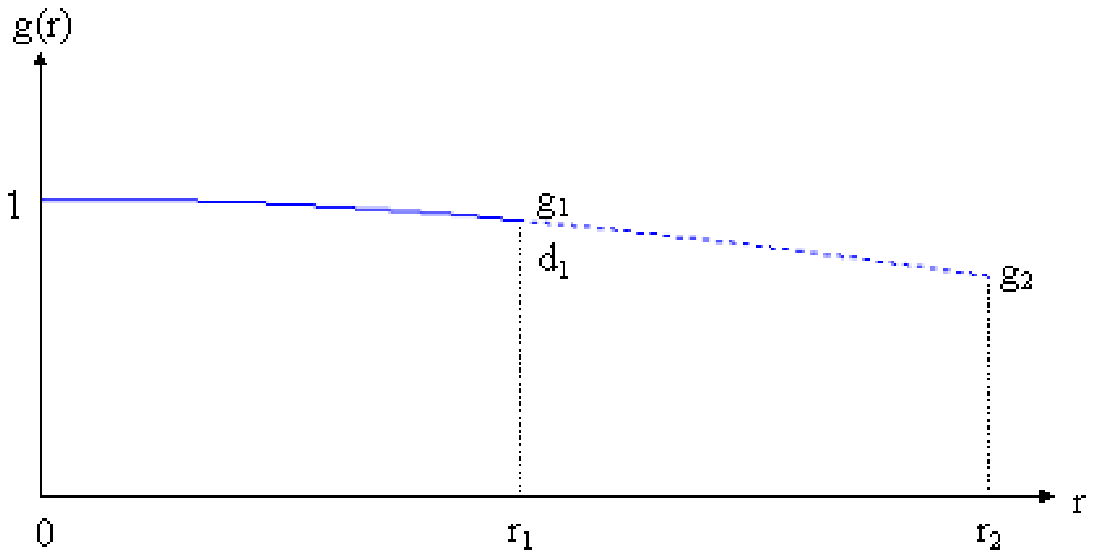}
\caption{A piecewise continuous function (two-segment).}
\label{fig: illustration}
\end{figure}

The appealing feature of having analytical geometric undistortion formulae when using the functions $\#5$ and $\#6$ in Table~\ref{table: 10 models} may come with a price. The simple model structure may limit the fitting flexibility and hence the fitting accuracy. In this case, a piecewise fitting idea can be applied to enhance accuracy of the simplified geometric distortion modeling, which is illustrated in Fig.~\ref{fig: illustration} with two segments. 

When using the function $\#5$ in Table~\ref{table: 10 models} for each segment of $f(r, {\bf k}_1)$ or $f(r, {\bf k}_2)$, the two segments are of the form
\begin{equation}
\label{eqn: piecewise 5}
\left\{
\begin{array}{l}
\displaystyle g_1(r) = \frac{1}{1 + {\bar k}_1 \, r}, \;\;\; {\rm for} \;\; r \; \in [0, r_1]\\[5pt]
\displaystyle g_2(r) = \frac{1}{a + {\bar k}_2 \, r}, \;\;\; {\rm for} \;\; r \; \in (r_1, r_2]
\end{array}\right.\hspace{-1mm},
\end{equation}
with $r_1=r_2/2$. To ensure that the overall function (\ref{eqn: piecewise 5}) is continuous across the interior knot, the following 3 constraints can be applied
\begin{equation}
\label{eqn: 3 constraints}
\displaystyle \frac{1}{1 + {\bar k}_1 \, r_1} = g_1, \;\;\frac{1}{a + {\bar k}_2 \, r_1} = g_1, \;\; \frac{1}{a + {\bar k}_2 \, r_2} = g_2, 
\end{equation}
where $g_1 = g_1(r_1) = g_2(r_1)$ and $g_2 = g_2(r_2)$. Since the coefficients $({\bar k}_1, a, {\bar k}_2)$ can be calculated from (\ref{eqn: 3 constraints}) uniquely by
\begin{equation}
\label{eqn: 3 solutions}
\left \{\hspace{-1mm}
\begin{array}{l}
{\bar k}_1 = (1/g_1 - 1)/r_1			\\
{\bar k}_2 = (1/g_2 - 1/g_1)/(r_2-r_1)	\\
a = 1/g_1 - {\bar k}_2 \, r_1
\end{array}\right.,
\end{equation}
the geometric distortion coefficients that are used in the nonlinear optimization can be chosen to be $(g_1, g_2)$ with the initial values $(1,1)$. During the nonlinear optimization process, $({\bar k}_1, a, {\bar k}_2)$ are calculated from $(g_1, g_2)$ in each iteration. When using the function $\#6$ in Table~\ref{table: 10 models}, similar functions to (\ref{eqn: 3 constraints}) and (\ref{eqn: 3 solutions}) can be derived by substituting $(r_1, r_2)$ with $(r_1^2, r_2^2)$. Furthermore, the piecewise idea can be easily extended to more segments. 

When applying the piecewise idea using the functions $\#5$ and $\#6$ in Table~\ref{table: 10 models}, better calibration accuracy can be achieved yet the property of having analytical geometric undistortion formulae can be retained. The   the above feature is clearly at the expense of more segments, i.e., more distortion coefficients to be   searched in the optimization process.

\section{Experimental Results and Discussions}
\label{sec: experimental results}

\subsection{Comparison Between the Simplified Geometric and the Radial Distortion Models}

\begin{figure}[htb]
\centering
\includegraphics[width=0.95\textwidth]{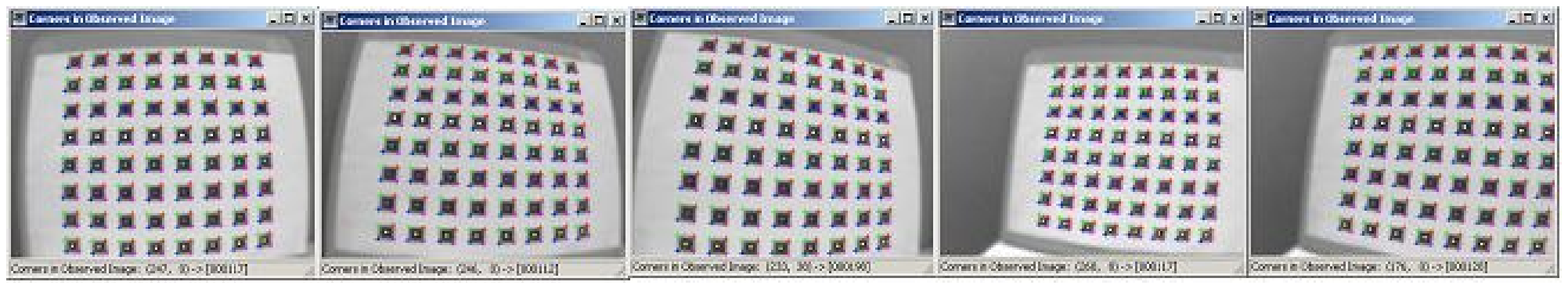}
\caption {Five images of the model plane with the extracted corners (indicated by cross) for the desktop camera.}
\label{fig: extracted desktop}
\end{figure}

\begin{figure}[htb]
\centering
\includegraphics[width=0.95\textwidth]{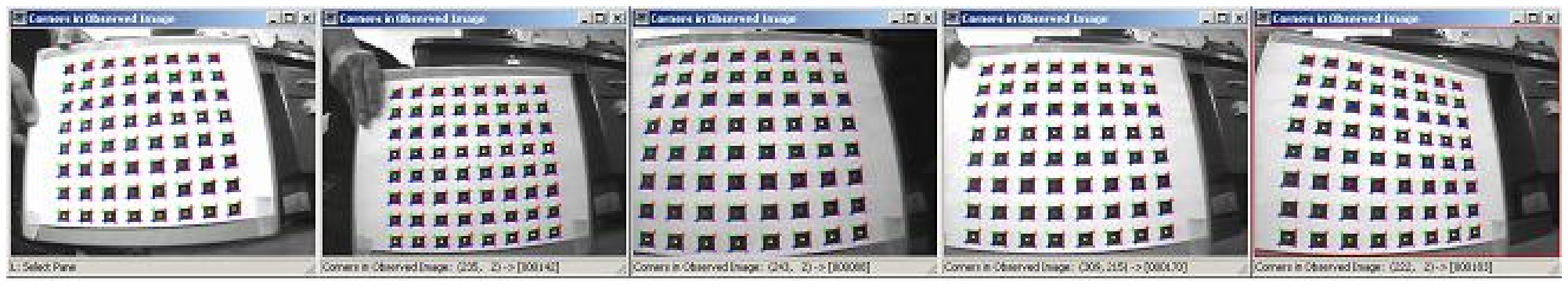}
\caption {Five images of the model plane with the extracted corners (indicated by cross) for the ODIS camera.}
\label{fig: extracted ODIS}
\end{figure}

Using three groups of test images (the public domain test images \cite{zhang98calibrationwebpage}, the desktop camera images \cite{Lilicalreport02} (a color camera), and the ODIS camera images \cite{Lilicalreport02,odiscamera} (the camera on the ODIS robot built at Utah State University \cite{lili02visualservo})), the final values of $J$ of the simplified geometric distortion model (\ref{eqn: U-D model in xdyd}) after nonlinear optimization by the Matlab function {\tt fminunc} using the 10 functions in Table~\ref{table: 10 models} are shown in Table~\ref{table: SGR comparisons}, where the values of $J$ using the same function but under the assumption of the radial distortion are also listed for comparison. In Table~\ref{table: SGR comparisons}, the numbers 1-10 in the first column denote the 10 functions in Table~\ref{table: 10 models} in the same order. 
The extracted corners for the model plane of the desktop and the ODIS cameras are shown in Figs.~\ref{fig: extracted desktop} and \ref{fig: extracted ODIS}, where the plotted dots in the center of each square are only used for judging the correspondence with the world reference points. 

\begin{table*}[htb]
\centering
\caption{Objective Function $J$ of The Simplified Geometric and The Radial Distortions using the 10 Functions in Table~\ref{table: 10 models}}
\label{table: SGR comparisons}
\renewcommand{\arraystretch}{1.1}
\vspace{-2 mm}
\setlength{\tabcolsep}{1.8mm}
{\small
{\begin {tabular}{|c||r|r||r|r||r|r|}\hline
&\multicolumn{2}{|c||}{\bf Public Images}&\multicolumn{2}{|c||}{\bf Desktop Images}& \multicolumn{2}{|c|}{\bf ODIS Images}\\\cline{2-7}
\raisebox{1.8ex}[0pt]{$\#$}&\multicolumn{1}{|r|}{Geometric} &\multicolumn{1}{|r||}{Radial}&\multicolumn{1}{|r|}{Geometric}&\multicolumn{1}{|r||}{Radial}&\multicolumn{1}{|r|}{Geometric}&\multicolumn{1}{|r|}{Radial}\\\hline
1 &  180.4617 & 180.5713 &  999.6644 & 1016.7437 &  928.9073 &  944.4418 \\\hline
2 &  148.2608 & 148.2788 &  904.0705 &  904.6796 &  913.1676 &  933.0981 \\\hline
3 &  145.5766 & 145.6592 &  801.3148 &  803.3074 &  836.9277 &  851.2619 \\\hline
4 &  144.8226 & 144.8802 &  777.3812 &  778.9767 &  825.8771 &  840.2649 \\\hline
5 &  184.9429 & 185.0628 & 1175.7494 & 1201.8001 & 1019.8750 & 1036.6208 \\\hline
6 &  146.9811 & 146.9999 &  797.9312 &  798.5720 &  851.6244 &  867.6192 \\\hline
7 &  145.3864 & 145.4682 &  786.2204 &  787.6185 &  830.6345 &  845.0206 \\\hline
8 &  145.3688 & 145.4504 &  784.8960 &  786.3590 &  829.4675 &  843.7991 \\\hline
9 &  144.7560 & 144.8328 &  779.0693 &  780.9060 &  823.0736 &  837.9181 \\\hline
10&  144.7500 & 144.8256 &  777.9869 &  780.0391 &  823.2726 &  838.3245 \\\hline\hline
  &           & 144.8179 &           &  776.7103 &           &  837.7749 \\\hline
\end {tabular}}}
\end{table*}

From Table~\ref{table: SGR comparisons}, the values of $J$ of the simplified geometric distortion models are generally smaller than those of the radial distortion models. The improvement for the public and desktop cameras are not significant, while it is significant for the ODIS camera. However, the above comparison between the simplified geometric and the radial distortion models might not be fair since the geometric models have more coefficients and it is evident that each additional coefficient in the model tends to decrease the fitting residual. Due to this concern, the objective function $J$ of the radial distortion modeling using 6 coefficients in equation (\ref{eqn: general polynomial}) ($2\times$ the maximal number of coefficients used in the geometric modeling) for the three groups of test images are also shown in the last row of Table~\ref{table: SGR comparisons}. 
The reason for choosing the radial distortion model (\ref{eqn: general polynomial}) with 6 coefficients is that this model is conventionally used and it is always among the best models in Table~\ref{table: 10 models} giving the top performance. 
Again, for the ODIS images, it is observed that the values of $J$ of the geometric modeling are all smaller than that of the radial distortion modeling using 6 coefficients in (\ref{eqn: general polynomial}), where the 6 coefficients are $(-0.3601,0.1801,-0.5149,3.1911,-6.4699,4.1625)$, except for the functions $\# 1,2,5,6$, which are relatively simple in complexity and have fewer distortion coefficients. It can thus be concluded that the distortion of the ODIS camera is not as perfectly radially symmetric as the other two cameras and the geometric modeling is more appropriate for the ODIS camera. 

\begin{table*}[htb]
\centering
\caption{Comparisons Between The Simplified Geometric and The Radial Distortion Models for the ODIS Images}
\label{table: Results using ODIS Images}
\renewcommand{\arraystretch}{1.1}
\setlength{\tabcolsep}{1.4mm}
\vspace{-2mm}
{\footnotesize
{\begin {tabular}{|c|c|r|ccc|ccc|ccccc|}\hline
{\bf Distortion}& $\#$& \multicolumn{1}{|c|}{$J$} &\multicolumn{3}{|c|}{${\bf k}_1$}&\multicolumn{3}{|c|}{${\bf k}_2$} 
& $\alpha$ & $\gamma$ & $u_0$& $\beta$&$v_0$ \\\hline
                &1  & 928.9073 & -0.2232 &      -&       -&  -0.2413&       -&        -&   272.5073& -0.0784&  140.7238&  268.7688&  115.5717\\\cline{2-14}
                &2  & 913.1676 & -0.2624 &      -&       -&  -0.2890&       -&        -&   256.7545& -0.4848&  137.3176&  252.4421&  117.6516\\\cline{2-14}
                &3  & 836.9277 & -0.1150 &-0.1305&       -&  -0.1206& -0.1454&        -&   264.5867& -0.3322&  140.4929&  260.2474&  115.0102\\\cline{2-14}
{\bf Simplified}&4  & 825.8771 & -0.3386 & 0.1512&       -&  -0.3718&  0.1756&        -&   259.4480& -0.2434&  140.5699&  255.2091&  114.8777\\\cline{2-14}
{\bf Geometric} &5  &1019.8750 &  0.2679 &      -&       -&   0.2968&       -&        -&   275.9477& -0.0049&  139.6337&  272.7017&  117.0080\\\cline{2-14}
{\bf Distortion}&6  & 851.6244 &  0.3039 &      -&       -&   0.3348&       -&        -&   258.0766& -0.3970&  139.5523&  253.7985&  115.7800\\\cline{2-14}
                &7  & 830.6345 & -0.0826 & 0.1964&       -&  -0.0768&  0.2320&        -&   263.1308& -0.3143&  140.6762&  258.3793&  114.9656\\\cline{2-14}
                &8  & 829.4675 &  0.0736 & 0.2259&       -&   0.0685&  0.2608&        -&   262.8587& -0.3068&  140.7462&  258.1623&  114.9220\\\cline{2-14}
                &9  & 823.0736 &  0.9087 & 0.8695&  0.5494&   1.6571&  1.4811&   0.8974&   259.5748& -0.2509&  140.9331&  251.9627&  114.7501\\\cline{2-14}
                &10 & 823.2726 &  0.2719 & 0.0232&  0.5950&   0.6543& -0.0563&   1.1524&   260.8910& -0.2444&  140.8209&  253.8259&  114.8106\\\hline\hline
&1 &944.4418&-0.2327&      -&     -&-&-&-&  274.2660 &  -0.1153 & 140.3620 & 268.3070 & 114.3916\\\cline{2-14}
&2 &933.0981&-0.2752&      -&     -&-&-&-&  258.3193 &  -0.5165 & 137.2150 & 252.6856 & 115.9302\\\cline{2-14}
&3 &851.2619&-0.1192&-0.1365&     -&-&-&-&  266.0850 &  -0.3677 & 139.9198 & 260.3133 & 113.2412\\\cline{2-14}
&4 &840.2649&-0.3554& 0.1633&     -&-&-&-&  260.7658 &  -0.2741 & 140.0581 & 255.1489 & 113.1727\\\cline{2-14}
{\bf Radial}&5&1036.6208& 0.2828&      -&     -&-&-&-&  278.0218 &  -0.0289 & 139.5948 & 271.9274 & 116.2992\\\cline{2-14}
{\bf Distortion}&6 &867.6192& 0.3190&      -&     -&-&-&-&  259.4947 &  -0.4301 & 139.1252 & 253.8698 & 113.9611\\\cline{2-14}
&7 &845.0206&-0.0815& 0.2119&     -&-&-&-&  264.4038 &  -0.3505 & 140.0528 & 258.6809 & 113.1445\\\cline{2-14}
&8 &843.7991& 0.0725& 0.2419&     -&-&-&-&  264.1341 &  -0.3429 & 140.1092 & 258.4206 & 113.1129\\\cline{2-14}
&9 &837.9181& 1.2859& 1.1839&0.7187&-&-&-&  259.2880 &  -0.2824 & 140.2936 & 253.7043 & 113.0078\\\cline{2-14}
&10&838.3245& 0.4494&-0.0124&0.8540&-&-&-&  260.9370 &  -0.2804 & 140.2437 & 255.3178 & 113.0561\\\hline
\end {tabular}}}
\end{table*}

\begin{figure}[htb]
\centering
\includegraphics[width=0.6\textwidth]{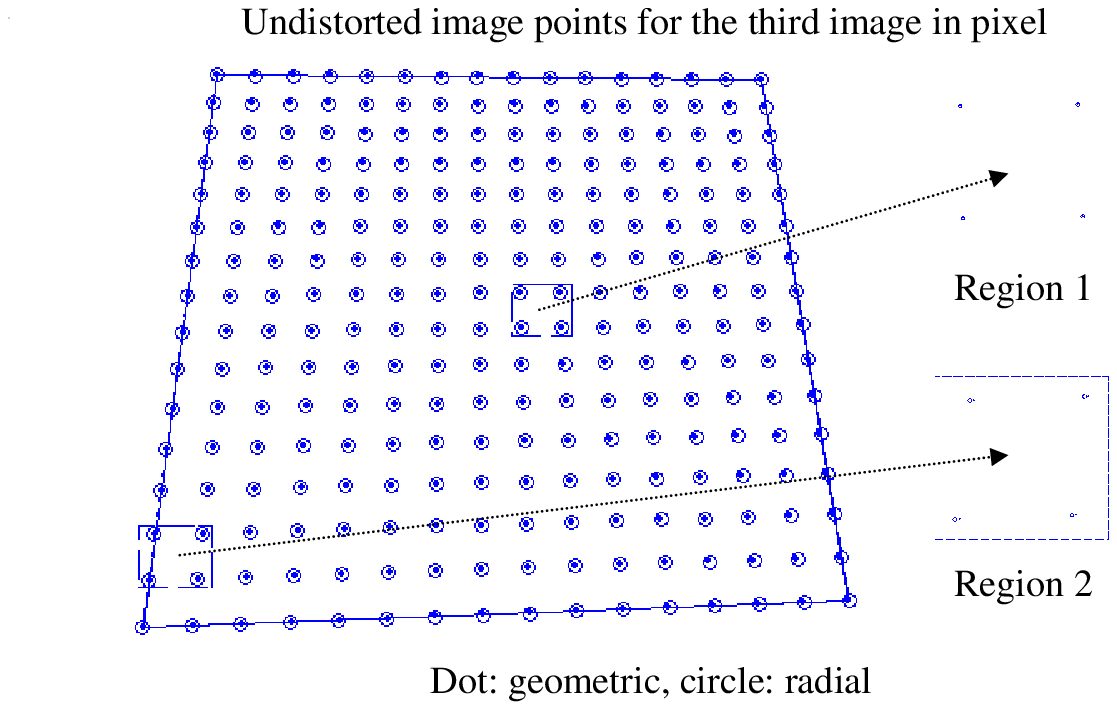}
\caption {Undistorted image points for the third image in Fig.~\ref{fig: extracted ODIS} using the simplified geometric distortion model $\#4$ and the radial distortion model (\ref{eqn: general polynomial}) with 6 coefficients.}
\label{fig: ODIS compare}
\end{figure}

\begin{figure}[htb]
\centering
\includegraphics[width=0.6\textwidth]{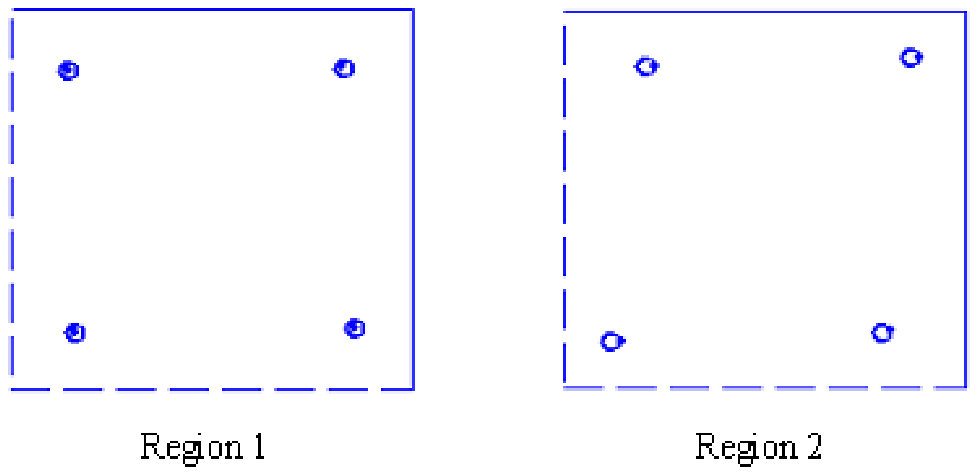}
\caption {Enlarged plot of the two regions in Fig.~\ref{fig: ODIS compare}.}
\label{fig: 2 regions}
\end{figure}

Figure~\ref{fig: ODIS compare} shows the undistorted image points for the third image in Fig.~\ref{fig: extracted ODIS} using the simplified geometric distortion model $\#4$ and the radial distortion model (\ref{eqn: general polynomial}) with 6 coefficients. The difference between the undistorted image points using the above two models can be observed at the image boundary (the enlarged plots of region$_1$ and region$_2$ are shown in Fig.~\ref{fig: 2 regions}), which is quite significant for applications that require sub-pixel accuracy.

\begin{figure}[htb]
\centering
\includegraphics[width=0.5\textwidth]{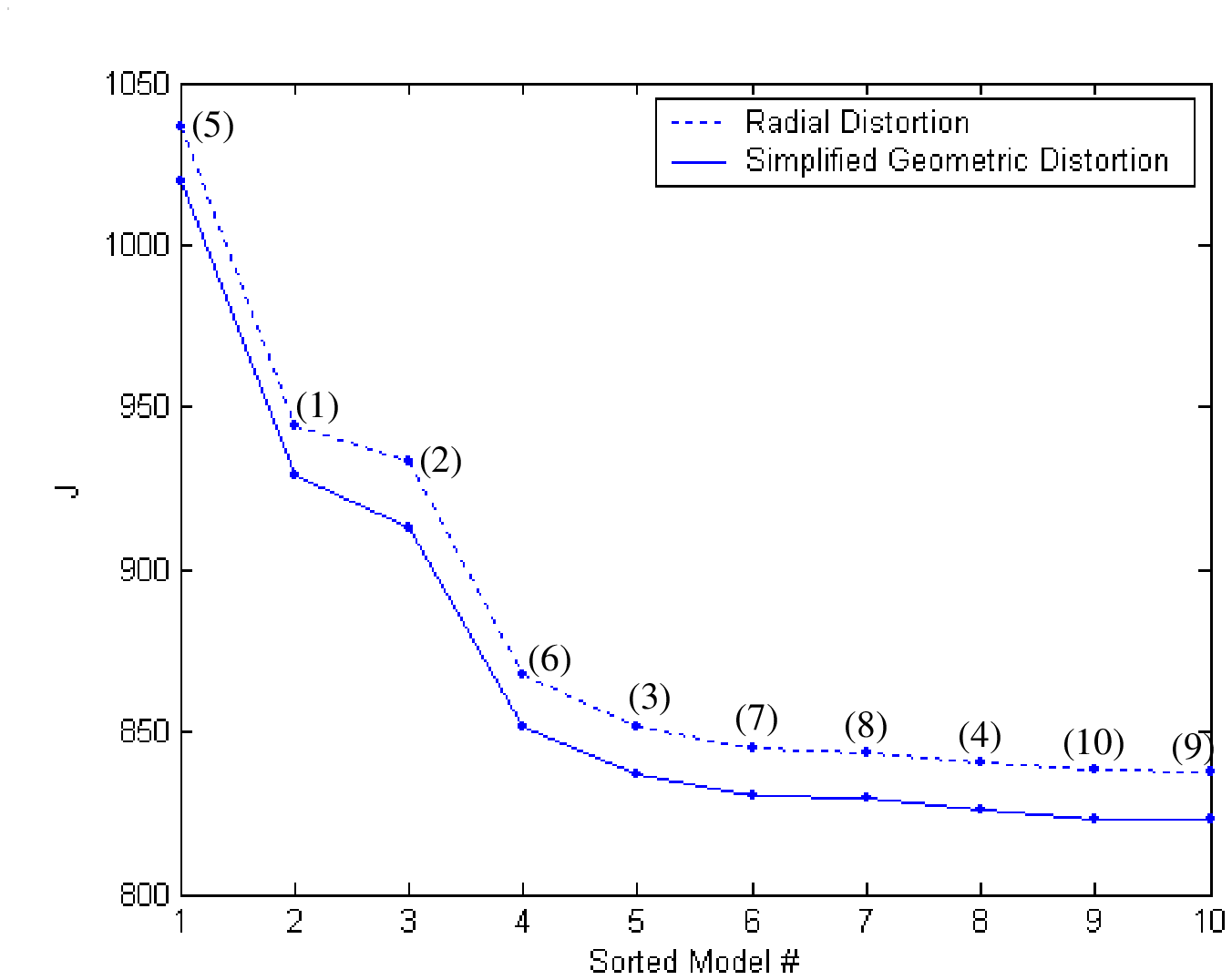}
\caption {Objective function $J$ of the simplified geometric and the radial distortion models for the ODIS images using the 10 functions in Table~\ref{table: 10 models} (corresponding model numbers are shown in the text).}
\label{fig: ODIS J s}
\end{figure}

The detailed estimated parameters using the 10 functions in Table~\ref{table: 10 models} for the simplified geometric and the radial distortion models are shown in Table~\ref{table: Results using ODIS Images} using the ODIS images, where the 5 intrinsic parameters are also listed for showing the consistency. Furthermore, the values of $J$ of the radial and geometric distortion models are plotted in Fig.~\ref{fig: ODIS J s} for the ODIS images, where the $x$ axis denotes the sorted model numbers in Table~\ref{table: 10 models} in an order with $J$ decreasing monotonously. 

\begin{figure}[htb]
\centering
\includegraphics[width=0.5\textwidth]{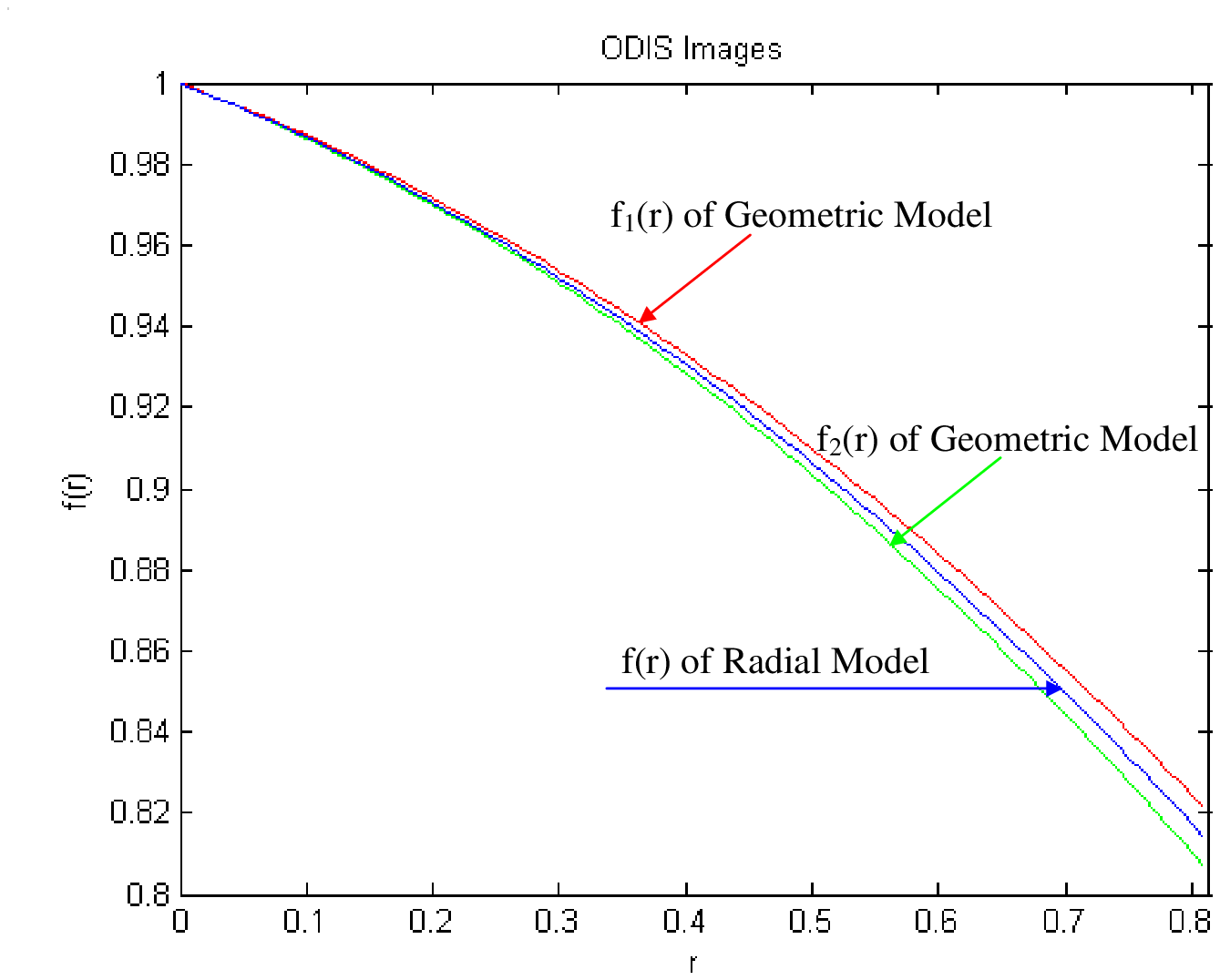}
\caption {$f(r, {\bf k}_1)$ and $f(r, {\bf k}_2)$ vs. $f(r, {\bf k})$ for the ODIS images using Model$_3$.}
\label{fig: ODIS f(r) s}
\end{figure}

From Table~\ref{table: Results using ODIS Images}, it is observed that the distortion coefficients ${\bf k}$ for the radial distortion models always lie between the corresponding values of ${\bf k}_1$ and ${\bf k}_2$ for the simplified geometric distortion models. Due to this reason, the resultant $f(r, {\bf k})$ curves also lie between the $f(r, {\bf k}_1)$ and $f(r, {\bf k}_2)$ curves, which can be seen from Fig.~\ref{fig: ODIS f(r) s}, where the $f(r, {\bf k})$, $f(r, {\bf k}_1)$, and $f(r, {\bf k}_2)$ curves for the ODIS images using the third function in Table~\ref{table: 10 models} (referred to as Model$_3$ hereafter) are plotted as an example. It can thus be concluded that when using one $f(r, {\bf k})$, it tries to find a compromise between $f(r, {\bf k}_1)$ and $f(r, {\bf k}_2)$. 

\begin{figure}[htb]
\centering
\includegraphics[width=0.5\textwidth]{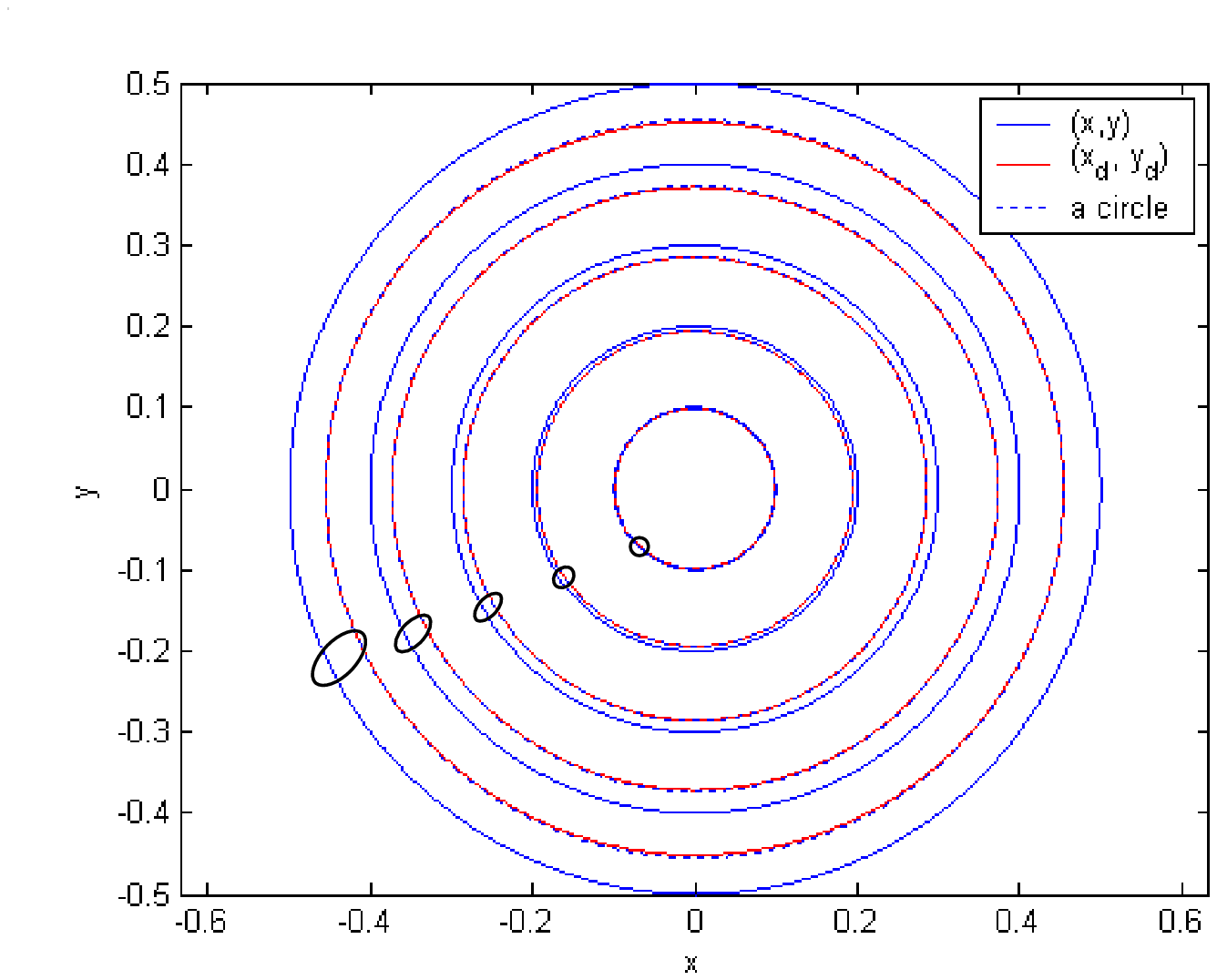}
\caption{$(x_d, y_d)$ vs. $(x, y)$ for the ODIS images under the simplified geometric distortion assumption using Model$_3$.}
\label{fig: ODIS ellipse}
\end{figure}

From Fig.~\ref{fig: ODIS f(r) s}, the difference between $f(r, {\bf k}_1)$ and $f(r, {\bf k}_2)$ increases as $r$ increases, which is barely noticeable at $r = 0.1$ but begins to be observable at $r =  0.3$. This information can also be seen from Fig.~\ref{fig: ODIS ellipse}, where the relationship between $(x,y)$ and $(x_d, y_d)$ is plotted for the ODIS images using Model$_3$. When using two different functions to model the distortion along the two image axes, the distortion shown in Fig.~\ref{fig: ODIS ellipse} is not exactly a smaller circle (since ${\bf k_1} < {\bf 0}$ and ${\bf k}_2 < {\bf 0}$), but a wide ellipse that is slightly shorter in the $y$ direction.

\begin{remark}
Classical criteria that are used in the computer vision to assess the accuracy of calibration includes the radial distortion \cite{Juyang92distortionmodel}. However, to our best knowledge, there is not a systematically quantitative and universally accepted criterion in the literature for performance comparisons among different distortion models. Due to this lack of criterion, in our work, the comparison is based on, but not restricted to, the fitting residual of the full-scale nonlinear optimization in (\ref{eqn: objective function}). 
\end{remark}

\subsection{Comparison Between the Simplified Geometric and the Piecewise Geometric Distortion Models using Functions $\#5$ and $\#6$ in Table~\ref{table: 10 models}}
\label{sec: piecewise and simplified}

Using the ODIS images, the final values of $J$ for the 1-segment, 2-segment, and 3-segment piecewise rational geometric distortion models using the functions $\#5$ and $\#6$ in Table~\ref{table: 10 models} are shown in Table~\ref{table: Results using ODIS Images piecewise}, where $f(r, {\bf k}_1)$ and $f(r, {\bf k}_2)$ have 1, 2, or 3 components depending on the number of segments used. The maximal range of $r$ is listed in the last column for each case. From Table~\ref{table: Results using ODIS Images piecewise}, it is observed that the values of $J$ after applying the piecewise idea are always smaller than those using fewer segments. A careful comparison between the values of $J$ of the 3-segment piecewise geometric distortion modeling using the function $\#6$ in Table~\ref{table: Results using ODIS Images piecewise} and the simplified geometric distortion models in Table~\ref{table: SGR comparisons} shows that the 3-segment piecewise geometric modeling using the function $\#6$ can have fairly good results. The piecewise idea is thus more suitable for applications that require real time image undistortion. 

\begin{table*}[htb]
\centering
\caption{Comparisons Between The Geometric and The Piecewise Geometric Distortion Models}
\label{table: Results using ODIS Images piecewise}
\renewcommand{\arraystretch}{1.05}
\setlength{\tabcolsep}{1.8mm}
\vspace{-2mm}
{\small
{\begin {tabular}{|c|c|c|r|ccc|ccc|c|}\hline
{\bf Distortion}&\bf Images&$\#$& \multicolumn{1}{|c|}{$J$} &\multicolumn{3}{|c|}{\bf $f(r, {\bf k}_1)$ Values}&\multicolumn{3}{|c|}{\bf $f(r, {\bf k}_2)$ Values} & $r_{\rm max}$ \\\hline
          &                                &5&147.8709 &0.9960&0.9830&0.9651&0.9957&0.9825&0.9646&0.4252\\\cline{3-11}
\bf 3-Segment&\raisebox{1.5ex}[0pt]{Public}&6&\bf 144.9397 &0.9957&0.9830&0.9646&0.9952&0.9825&0.9640&0.4260\\\cline{2-11}
\bf Geometric&                             &5&802.8802 &0.9699&0.8954&0.8047&0.9666&0.8957&0.8048&0.8643\\\cline{3-11}
\bf Distortion&\raisebox{1.5ex}[0pt]{Desktop}&6&\bf 782.3082 &0.9726&0.8980&0.8069&0.9712&0.8991&0.8088&0.8673\\\cline{2-11}
          &                                &5&840.2963 &0.9657&0.9035&0.8284&0.9720&0.9022&0.8234&0.8167\\\cline{3-11}
          &\raisebox{1.5ex}[0pt]{ODIS}     &6&\bf 823.8911 &0.9709&0.9084&0.8327&0.9741&0.9044&0.8253&0.8182\\\hline\hline          
          &                                &5&149.5355 &0.9871&0.9630&     -&0.9862&0.9622&     -&0.4250\\\cline{3-11}
\bf 2-Segment &\raisebox{1.5ex}[0pt]{Public}&6&145.7634 &0.9902&0.9639&     -&0.9895&0.9634&     -&0.4263\\\cline{2-11}
\bf Geometric &                            &5&824.7348 &0.9167&0.7949&     -&0.9168&0.7996&     -&0.8612\\\cline{3-11}
\bf Distortion&\raisebox{1.5ex}[0pt]{Desktop}&6&787.4081 &0.9380&0.8027&     -&0.9391&0.8064&     -&0.8665\\\cline{2-11}
          &                                &5&850.6158 &0.9249&0.8242&     -&0.9197&0.8107&     -&0.8101\\\cline{3-11}
          &\raisebox{1.5ex}[0pt]{ODIS}     &6&830.7617 &0.9449&0.8313&     -&0.9405&0.8182&     -&0.8176\\\hline\hline
          &                                &5& 184.9428&0.9587&     -&     -&0.9580&     -&     -&0.4216\\\cline{3-11}
\bf Single    &\raisebox{1.5ex}[0pt]{Public}&6& 146.9811&0.9642&     -&     -&0.9640&     -&     -&0.4264\\\cline{2-11}
\bf Geometric &                            &5&1175.6975&0.7983&     -&     -&0.8149&     -&     -&0.7968\\\cline{3-11}
\bf Distortion&\raisebox{1.5ex}[0pt]{Desktop}&6& 797.9354&0.8033&     -&     -&0.8057&     -&     -&0.8654\\\cline{2-11}         
          &                                &5&1019.8751&0.8261&     -&     -&0.8109&     -&     -&0.7852\\\cline{3-11}
          &\raisebox{1.5ex}[0pt]{ODIS}     &6& 851.6577&0.8331&     -&     -&0.8191&     -&     -&0.8119\\\hline
\end {tabular}}}
\end{table*}

\begin{figure}[htb]
\centering
\includegraphics[width=0.5\textwidth]{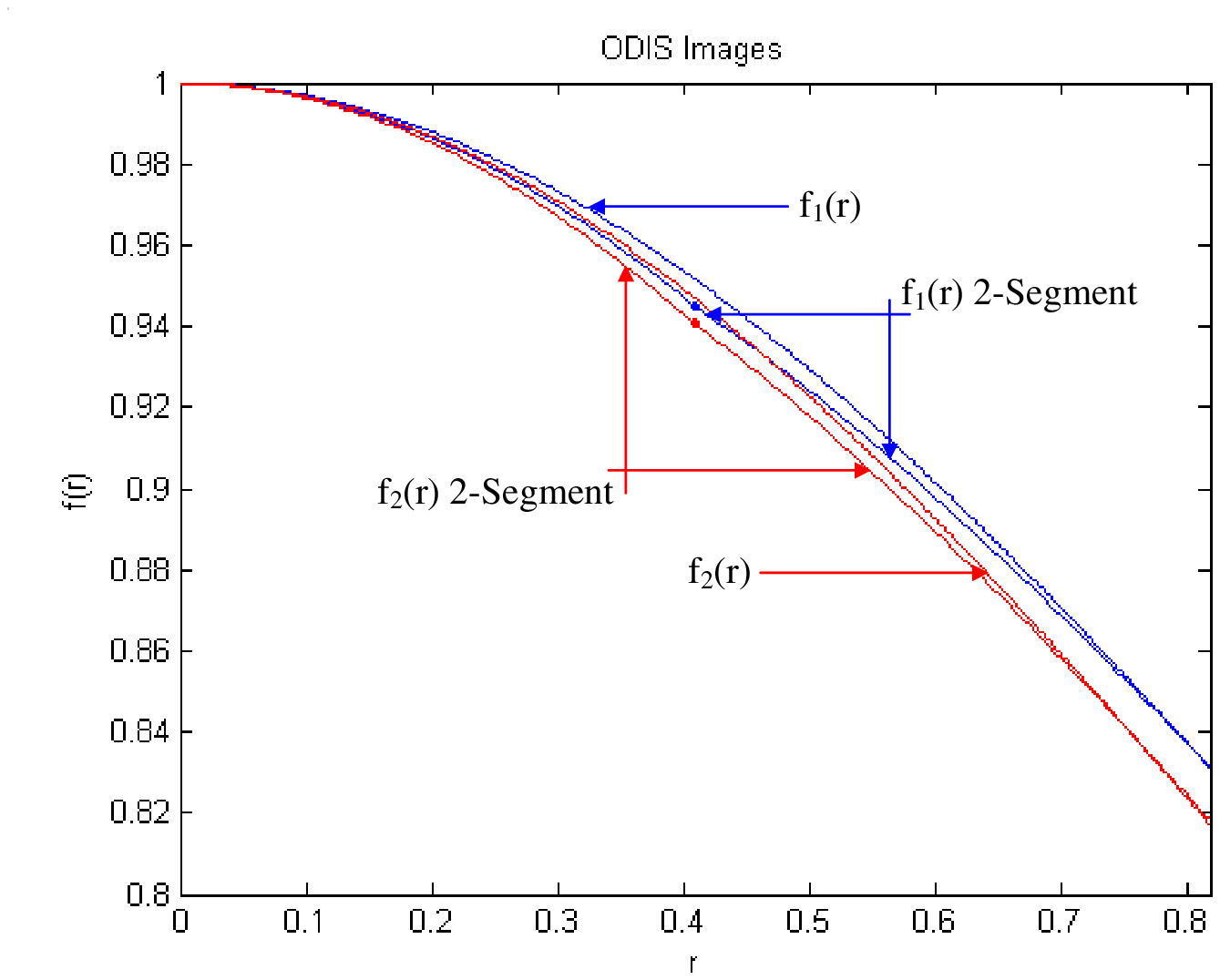}
\caption{$f(r)$ curves of the 2-segment and the single rational geometric distortion models using function $\#6$ in Table~\ref{table: 10 models} for the ODIS images.}
\label{fig: ODIS frs 2seg}
\end{figure}

\begin{figure}[htb]
\centering
\includegraphics[width=0.5\textwidth]{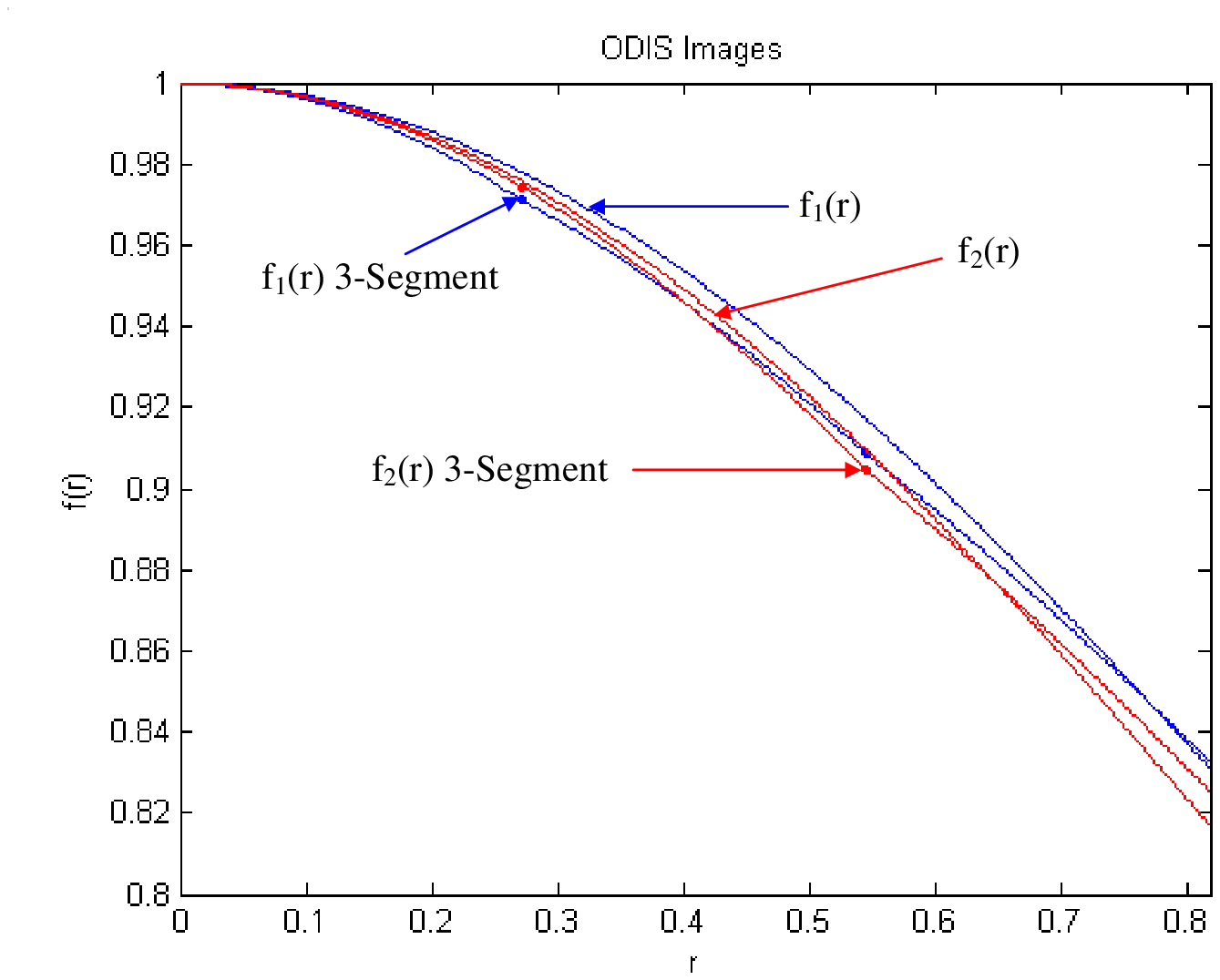}
\caption{$f(r)$ curves of the 3-segment and the single rational geometric distortion models using function $\#6$ in Table~\ref{table: 10 models} for the ODIS images.}
\label{fig: ODIS frs 3seg}
\end{figure}

The resulting estimated $f(r, {\bf k}_1)$ and $f(r, {\bf k}_2)$ curves of the 2-segment and 3-segment geometric distortion models using the rational function $\#6$ in Table~\ref{table: 10 models} for the ODIS images are plotted in Figs.~\ref{fig: ODIS frs 2seg} and \ref{fig: ODIS frs 3seg}. 

One issue in the implementation of the piecewise idea is how to decide $r_{\rm max}$, which is related to the estimated extrinsic parameters that are changing from iteration to iteration during the nonlinear optimization process. In our implementation, for each camera, 5 images are taken. $r_{\rm max}$ is chosen to be the maximum $r$ of all the extracted feature points on the 5 images for each iteration.

\subsection{Comparison Between the Geometric Modeling Methods (\ref{eqn: U-D model in xdyd}) and (\ref{eqn: geometric using circular control points})}

Both using 6 coefficients, the values of $J$ of the simplified geometric distortion modeling method (\ref{eqn: U-D model in xdyd}) using function $f(r) = 1 + k_1 r^2 + k_2 r^4 + k_3 r^6$ are shown in Table~\ref{table: geometrics} for the three groups of test images, where $J$ of the geometric modeling method (\ref{eqn: geometric using circular control points}) in \cite{Janne00Geometric} with the distortion coefficients $(k_1, k_2, k_3, p_1, p_2, p_3)$ is also listed for comparison. From Table~\ref{table: geometrics}, it is observed that the simplified geometric modeling method, though simpler in structure, does not necessarily give a less accurate calibration performance. 

\begin{table}[htb]
\centering
\caption{Comparison Between The Geometric Distortion Modeling Methods (\ref{eqn: U-D model in xdyd}) and (\ref{eqn: geometric using circular control points})}
\label{table: geometrics}
\renewcommand{\arraystretch}{1}
\setlength{\tabcolsep}{1.8mm}
\vspace{-2mm}
{\small
{\begin {tabular}{|c|c|c|c|}\hline
\bf Eqn. & \bf Public Images & \bf Desktop Images  & \bf ODIS Images\\\hline
(\ref{eqn: U-D model in xdyd})                       & 144.7596  & 775.9196 & 823.9299 \\\hline
(\ref{eqn: geometric using circular control points}) & 142.9723  & 772.8905 & 834.5090 \\\hline
\end {tabular}}}
\end{table}

\begin{remark}
To make the results in this paper reproducible by other researchers for further investigation, we present the options we use for the nonlinear optimization: \texttt{options = optimset(`Display', `iter', `LargeScale', `off', `MaxFunEvals', 8000, `TolX', $10^{-5}$,  `TolFun', $10^{-5}$, `MaxIter', 120)}. The raw data of the extracted feature locations in the image plane are also available \cite{Lilicalreport02}. 
\end{remark}

\section{Concluding Remarks}
\label{sec: conclusion}
In this paper, a family of simplified geometric distortion models are proposed that apply different polynomial and rational functions along the two image axes. Experimental results are presented to show that the proposed simplified geometric distortion modeling method can be more appropriate for cameras whose distortion is not perfectly radially symmetric around the center of distortion. Analytical geometric undistortion is possible using two of the distortion functions discussed in this paper and their performance can be improved by applying a piecewise idea. 

The proposed simplified geometric distortion modeling method is simpler than that in \cite{Janne00Geometric}, where the nonlinear geometric distortion is further classified into the radial distortion and the decentering distortion and the total distortion is a sum of these distortion effects. Though simple in the structure, the simplified geometric distortion modeling gives comparable performance to that in \cite{Janne00Geometric}. Furthermore, for some cameras, like the ODIS camera studied here, the simplified geometric distortion modeling can even perform better.

In this paper, we are restricting the maximal number of distortion coefficients considered to be 3 in all the distortion functions in Table~\ref{table: 10 models}, because it has also been found that too high an order may cause numerical instability \cite{tsai87AVersatile,zhang99calibrationinpaper,GWei94Implicit}. However, the appropriate number of distortion coefficients should not be determined only by a numerical issue. A stronger argument should come from the relationship between $J$ and the number of distortion coefficients. The appropriate number of distortion coefficients is chosen when the calibration accuracy does not show to have much improvement as the number of distortion coefficients increases beyond this value. 

The comparison between the piecewise and the simplified geometric distortion models in Sec.~\ref{sec: piecewise and simplified} brings up the question of preference between ``more segments with low-complexity function'' or ``more distortion coefficients with more complex function''. The above question is not answered in this work and is a direction of future investigation. 

\bibliography{calibration,csois1,csois2}
\end{document}